# Performance Comparison of Different Machine Learning Algorithms on the Prediction of Wind Turbine Power Generation

Onder Eyecioglu
Department of Computer Engineering, Faculty of Engineering and Architecture, Nisantasi University, Istanbul, Turkey.
oeyeci@gmail.com

Batuhan Hangun, Korhan Kayisli
Department of Electrical and Electronics Engineering, Faculty of Engineering and Architecture, Nisantasi University Istanbul, Turkey.
batuhanhangun@gmail.com
korhankayisli@gmail.com

Mehmet Yesilbudak
Department of Electrical and Electronics Engineering, Faculty of Engineering and Architecture, Nevsehir Haci Bektas Veli University, Nevsehir, Turkey.
myesilbudak@nevsehir.edu.tr

*Abstract*—Over the past decade, wind energy has gained more attention in the world. However, owing to its indirectness and volatility properties, wind power penetration has increased the difficulty and complexity in dispatching and planning of electric power systems. Therefore, it is needed to make the high-precision wind power prediction in order to balance the electrical power. For this purpose, in this study, the prediction performance of linear regression, k-nearest neighbor regression and decision tree regression algorithms is compared in detail. k-nearest neighbor regression algorithm provides lower coefficient of determination values, while decision tree regression algorithm produces lower mean absolute error values. In addition, the meteorological parameters of wind speed, wind direction, barometric pressure and air temperature are evaluated in terms of their importance on the wind power parameter. The biggest importance factor is achieved by wind speed parameter. In consequence, many useful assessments are made for wind power predictions.

## I. Introduction

People have been using fossil fuel-based energy supports for many years. However, population and energy requirements have also been increased, continuously. At this point, the limitations of fossil fuel reserves and their environmental damages have encouraged the people to find the alternative energy resources, such as solar, wind, hydroelectric, geothermal, biomass, hydrogen, wave, etc. [1-3]. Accordingly, researchers have been working about the effective and efficient usage of these resources. The assessment of renewable energy potential, development of new control techniques for electrical machines and usage of special control algorithms for power electronic devices are some of the main researches in this field [4-6]. Especially, many studies have been conducted in the literature in order to predict the wind turbine power generation.

*Liu et al.* combined Gaussian process regression and multiple imputation approaches, and handled wind power prediction with the missing values [7]. *Ouyang et al.* detected wind power ramp events with the Markov switching auto regression model and corrected the prediction residual of physical models [8]. *Naik et al.* hybridized variational mode decomposition and low rank multi-kernel ridge regression, and constructed the effective prediction intervals for wind power [9]. *He et al.* predicted the probability density based on quantile regression neural network and kernel density function, and quantified the uncertainties in wind power generation [10]. *Dowell et. al.* modelled the location parameter of logit-normal distribution as a sparse vector autoregressive process and improved the probabilistic prediction skill of wind power [11]. *Wang et al.* optimized the adaptive robust multi-kernel regression using variational Bayesian methods and obtained the deterministic and probabilistic prediction of wind power, simultaneously [12]. *Hu et al.* developed a support vector regression model based on the Gaussian noise with heteroscedasticity and showed the effectiveness of the developed model in wind power prediction [13]. *Yampikulsakul et al.* proposed a weighted version of least squares support vector regression for wind power systems and modelled the wind turbine response as a function of weather variables, efficiently [14]. *Xiyun et al.* used the Bootstrap quantile regression and found the upper and lower limits of the confidence intervals of wind power intervals [15]. *Qureshi et al.* utilized deep neural networks as base-regressors and a meta-regressor along with transfer learning, and smoothed out the rapid transients in wind power prediction behavior [16]. *Demollia et al.* also compared least absolute shrinkage selector operator, k-nearest neighbor, extreme gradient boost, random forest and support vector regression algorithms for the prediction of wind power values. The accuracy values were found to be satisfying for different locations [17].

In this study, the performance of three machine learning algorithms called linear regression, k-nearest neighbor regression and decision tree regression is compared for the prediction of wind turbine power generation. In the prediction phase, the meteorological parameters of wind speed, wind direction, barometric pressure and air temperature are utilized, and their importance factors on the wind power parameter are also revealed. As a result of overall analyses made in this study, wind power predictions are obtained in an accurate manner.

## II. Methodology

Machine learning is a field of computer science, which allows one to interpret a dataset, to derive a meaning from it, and then to use the revealed knowledge in the problem required to be solved. In other words, machine learning is a method of programming a computer to optimize any given performance criterion using an example dataset or experience [18]. In this study, three machine learning algorithms called linear regression, k-nearest neighbor regression and decision tree regression are used to predict the wind turbine power generation.

### A. Linear Regression

In linear regression, it is assumed that a linear system defines the relationship between the dependent variable $y$ and $p$-length regressors $x$ for a given dataset $\{y_i, x_{i_1}, x_{i_2}, \ldots, x_{i_p}\}_{i=1}^{n}$ with $n$ statistical units [19]. The mathematical representation of multiple linear regression can be written as follows, where $\beta_0$ is the intercept, $\beta_1, \beta_2, \ldots, \beta_p$ are the slopes and $\varepsilon_i$ is the residual (error).

$$y_i = \beta_0 + \beta_1 x_{i_1} + \beta_2 x_{i_2} + \cdots + \beta_p x_{i_p} + \varepsilon_i \quad (1)$$

### B. k-Nearest Neighbor Regression

k-nearest neighbor algorithm is a non-parametric pattern recognition method, which can be used for both classification and regression. This algorithm depends on calculating the distance values and determining the nearest neighbors for a given data point [20]. In k-nearest neighbor regression, it is allowed to estimate continuous variables. Its process steps are given below:

1-Compute the distance between a selected sample and other samples

2-Make an order of samples considering the computed distances.

3-Based on the root mean squared error calculated, find an optimal number of k nearest neighbors utilizing cross validation.

4-Find the inverse distance weighted average value by using the k-nearest neighbors.

### C. Decision Tree Regression

Generally, a decision tree may be defined as a data structure which implements a hierarchical divide-and-conquer method. Except from its top element, each element of a decision tree has a parent element [21]. This method is also an efficient non-parametric approach to be used both for regression and classification. An example of a decision tree is shown in Figure 1.

For node $m$, $X_m$ is a subset of $X$ which reached node m. So, it is basically the set of all $x \in X$, which satisfies all of the conditions in the decision nodes on its path from the root to node $m$. It can be defined as below:

$$b_m(x) = \begin{cases} 1 & \text{if } x \in X_m, x \text{ that reaches a specific node } m \\ 0 & \text{otherwise} \end{cases} \quad (2)$$

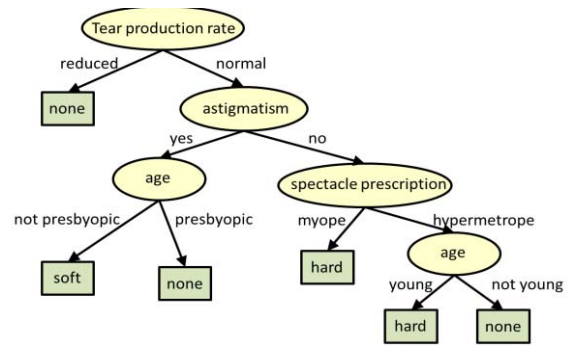

**Fig. 1.** An example of a decision tree [22]

### D. Error Measures

Mean absolute error (MAE) and coefficient of determination ($R^2$) measures are employed for comparing the prediction performance of linear, k-nearest neighbor and decision tree regressors in this study. These error measures are expressed as below, where $y_j$ and $\hat{y}_j$ are the actual response and predicted response of observation $j$ and $\bar{y}$ is the mean value of all actual responses [23, 24].

$$MAE = \frac{1}{n}\sum_{j=1}^{n}|y_j - \hat{y}_j| \quad (3)$$

$$R^2 = \frac{\sum_{j=1}^{n}(\hat{y}_j - \bar{y})^2}{\sum_{j=1}^{n}(y_j - \bar{y})^2} \quad (4)$$

## III. Results and Discussion

The dataset used in this study was obtained from [25]. It includes 4464 data points for wind speed (m/s), wind direction (°), barometric pressure (hPa), air temperature (°C) and wind power (kW) parameters recorded at 10-min intervals. The total dataset was divided into different parts in order to use them in training and testing steps. The statistical properties of the total dataset are also presented in Table I.

TABLE I. The Statistical Properties of the Total Dataset

| Properties | Features | | | | |
|---|---|---|---|---|---|
| | *Air Temperature* | *Barometric Pressure* | *Wind Direction* | *Wind Speed* | *Wind Power* |
| Mean Value | 3.9397 | 1019.464 | 243.1054 | 8.6540 | 666.60 |
| Standard Deviation | 2.0408 | 13.0539 | 55.1089 | 4.2409 | 716.68 |
| Minimum Value | -5.29 | 979.79 | 100.67 | 0.32 | 2.24 |
| Maximum Value | 10 | 1035.72 | 359.78 | 21.07 | 2033.12 |

The scatter plot of all features in the total dataset is illustrated in Figure 2. This scatter plot helps to deduct the correlation between each parameter. The diagonal and non-diagonals of this matrix also respectively show the variances of five random variables and the covariance between random variables. From this figure, it can be said that there is a positive correlation between wind speed and wind power features. The prediction results of linear, k-nearest neighbor and decision tree regressors are depicted in Figure 3 for a random day.

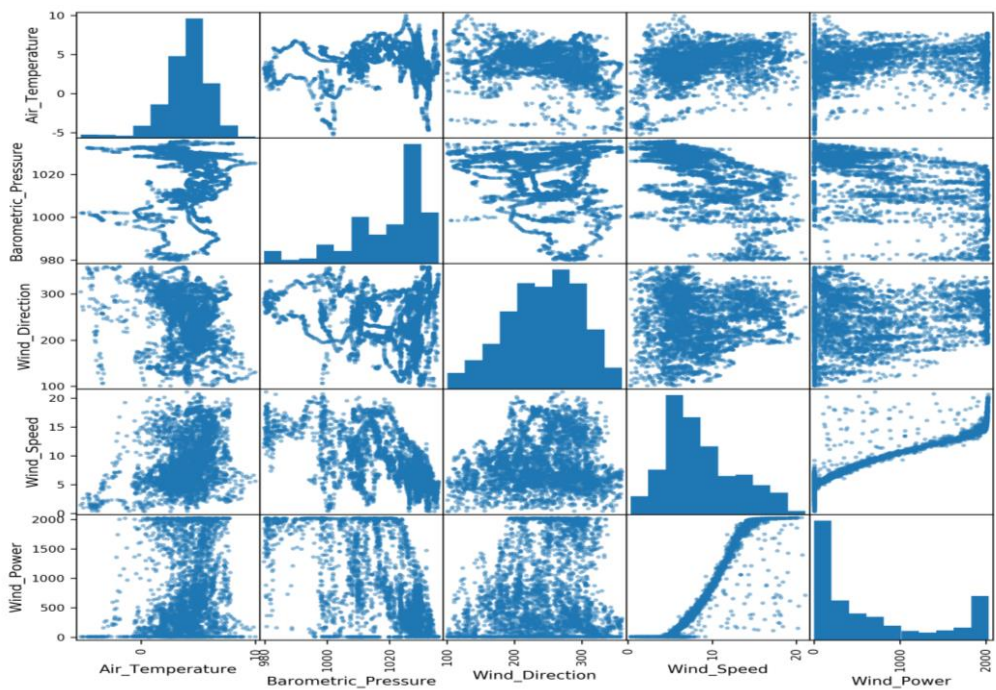

**Fig. 2.** The scatter plot of all features in the total dataset

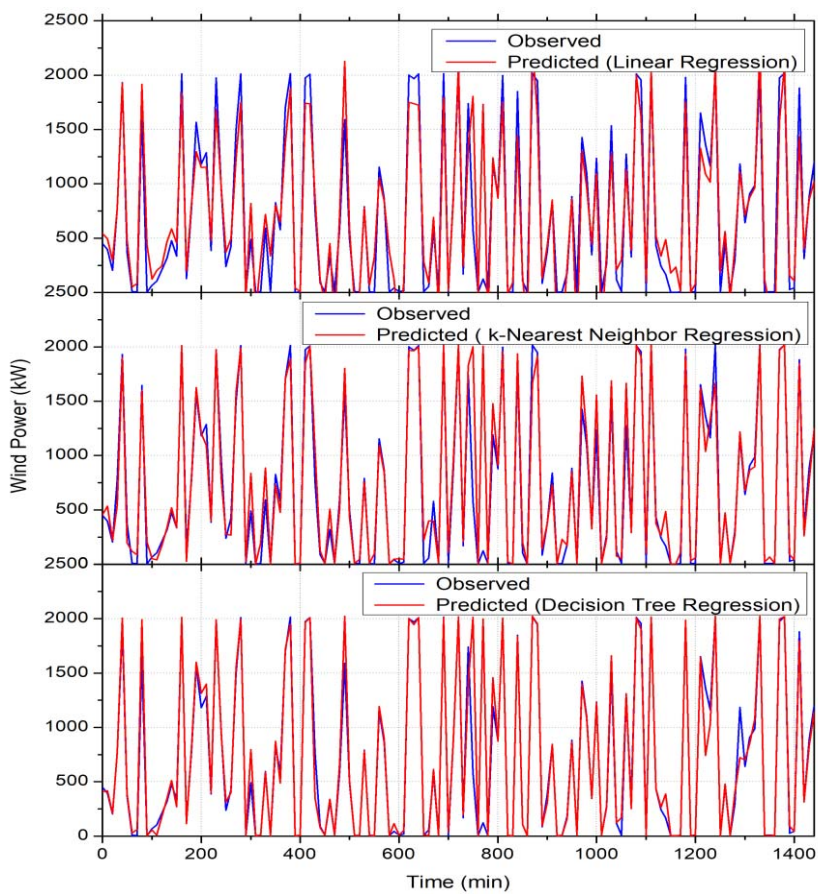

**Fig. 3.** The wind power prediction results of linear regression, k-nearest neighbor regression and decision tree regression

As seen in Figure 3, the wind power values predicted by linear regression, k-nearest neighbor regression and decision tree regression algorithms have good agreement with the observed ones. In addition to this figure, the statistical fit of actual and predicted wind power values in terms of coefficient of determination measure is shown in Figure 4 for the total test data.

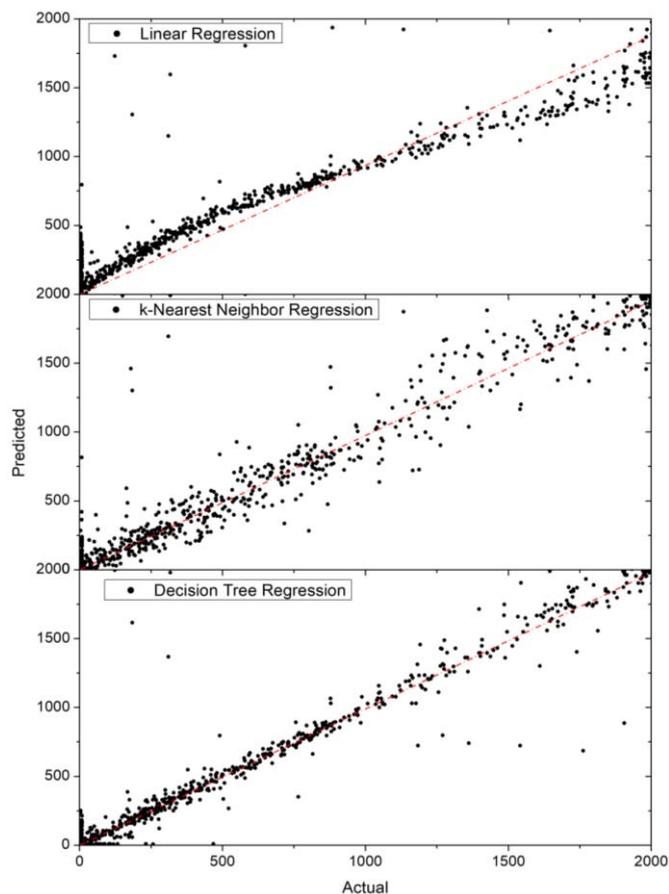

**Fig. 4.** The statistical fits of linear, k-nearest neighbor and decision tree regression models

The accuracy validations of the mentioned machine learning algorithms have been estimated by the k-fold cross validation procedure. It should be noted that the value of k has been chosen as 10. In this procedure, the input dataset is split into two parts as training data and testing data. 80% of all data samples are used as the training data, while the other remaining 20% of total dataset are utilized as the testing data. The accuracy values of the dependent variable of wind power in terms of each validation case are presented in Table II for the employed three regression algorithms.

As a result of the prediction process, mean absolute error and coefficient of determination values are listed in Table III, while the importance factors of the utilized meteorological features are provided in Table IV. In case of examining Tables III and IV in detail, on the one hand, in terms of the coefficient of determination results, k-nearest neighbor regression outperforms the other two regression models with achieving the $R^2$ of 0.94. On the other hand, in terms of the mean absolute error results, decision tree regression surpasses the other two regression models with accomplishing the MAE of 59.49 kW. Linear regression produces the worst prediction results with causing the MAE of 165.46 kW and $R^2$ of 0.88. In addition, among all of the utilized meteorological factors, wind speed parameter shows the biggest importance factor with the value of 0.8486.

TABLE II. THE ACCURACY VALUES OF K-FOLD CROSS VALIDATION METHOD

| k | Accuracy Values | | |
|---|---|---|---|
| | *Linear Regression* | *k-Nearest Neighbor Regression* | *Decision Tree Regression* |
| 0 | 0.8696 | 0.8778 | 0.8993 |
| 1 | 0.9092 | 0.9358 | 0.9191 |
| 2 | 0.8861 | 0.9262 | 0.8806 |
| 3 | 0.9127 | 0.9023 | 0.8905 |
| 4 | 0.8486 | 0.9043 | 0.9036 |
| 5 | 0.8461 | 0.9539 | 0.9578 |
| 6 | 0.9113 | 0.8968 | 0.8764 |
| 7 | 0.8876 | 0.9104 | 0.9299 |
| 8 | 0.9077 | 0.9140 | 0.8862 |
| 9 | 0.8668 | 0.9494 | 0.9326 |
| Average | 0.8846 | 0.9171 | 0.9076 |

TABLE III. ERROR RESULTS OF THE EMPLOYED MACHINE LEARNING ALGORITHMS

| *Algorithm* | *Error Results* | |
|---|---|---|
| | *MAE* | *$R^2$* |
| Linear Regression | 165.46 | 0.88 |
| k-Nearest Neighbor Regression | 93.13 | 0.94 |
| Decision Tree Regression | 59.49 | 0.91 |

TABLE IV. IMPORTANCE FACTORS OF THE UTILIZED METEOROLOGICAL FEATURES

| *Meteorological Features* | *Importance Factors* |
|---|---|
| Air Temperature | 0.0207 |
| Barometric Pressure | 0.0806 |
| Wind Direction | 0.0501 |
| Wind Speed | 0.8486 |

## IV. CONCLUSIONS

In this study, on the one hand, linear regression, k-nearest neighbor regression and decision tree regression algorithms are employed for comparing their prediction performance on the wind turbine power generation. In terms of the error measures of mean absolute error and coefficient of determination, k-nearest neighbor regression and decision tree regression algorithms provide better prediction performance than linear regression algorithm. On the other hand, wind speed, wind direction, barometric pressure and air temperature parameters are considered as the meteorological parameters affecting the wind turbine power generation. In terms of the importance factor achieved, wind speed parameter shows more promising impact than barometric pressure, wind direction and air temperature parameters. In future studies, more machine learning algorithms and more meteorological parameters should be taken into account for deeper analyses.

## REFERENCES

[1] R. Vieira, M.D. Bellar, J.V.S. Cunha, T.R. Oliveira and A.M.B. Aluisio, "Renewable energy system for small water desalination plant", 7th International Conference on Renewable Energy Research and Applications (ICRERA'18), pp. 1074-1079, 14-17 October 2018, Paris, France.

[2] A. Roy, F. Auger, S. Bourguet, F. Dupriez-Robin and Q.T. Tran, "Benefits of demand side management strategies for an island supplied by marine renewable energies", IEEE 7th International Conference on Renewable Energy Research and Applications (ICRERA'18), pp. 474-481, 14-17 October 2018, Paris, France.

[3] K. Amara, A. Fekik, D. Hocine, M.L. Bakir, E.B. Bourennane, T.A. Malek and A. Malek, "Improved performance of a PV solar panel with adaptive neuro fuzzy inference system ANFIS based MPPT", IEEE 7th International Conference on Renewable Energy Research and Applications (ICRERA'18), pp. 1098-1101, 14-17 October 2018, Paris, France.

[4] K. Anwar and S. Deshmukh, "Assessment and mapping of solar energy potential using artificial neural network and GIS technology in the southern part of India", International Journal of Renewable Energy Research, vol. 8, no. 2, pp. 974-985, June 2018.

[5] M. Benakcha, "Backstepping control of dual stator induction generator used in wind energy conversion system", International Journal of Renewable Energy Research, vol. 8, no. 1, pp. 384-395, March 2018.

[6] S. Das and A.K. Akella, "Power flow control of PV-wind-battery hybrid renewable energy systems for stand-alone application", International Journal of Renewable Energy Research, vol. 8, no. 1, pp. 36-43, March 2018.

[7] T. Liu, H. Wei and K. Zhang, "Wind power prediction with missing data using Gaussian process regression and multiple imputation", Applied Soft Computing, vol. 71, pp. 905-916, October 2018.

[8] T. Ouyang, X. Zha, L. Qin, Y. He and Z. Tang, "Prediction of wind power ramp events based on residual correction", Renewable Energy, vol. 136, pp. 781-792, June 2019.

[9] J. Naik, R. Bisoi and P.K. Dash, "Prediction interval forecasting of wind speed and wind power using modes decomposition based low rank multi-kernel ridge regression", Renewable Energy, vol. 129, pp. 357-383, December 2018.

[10] Y. He and H. Li, "Probability density forecasting of wind power using quantile regression neural network and kernel density estimation", Energy Conversion and Management, vol. 164, pp. 374-384, May 2018.

[11] J. Dowell and Pierre Pinson, "Very-short-term probabilistic wind power forecasts by sparse vector autoregression", IEEE Transactions on Smart Grid, vol. 7, no. 2, pp. 763-770, March 2016.

[12] Y. Wang, Q. Hu, D. Meng and P. Zhu, "Deterministic and probabilistic wind power forecasting using a variational Bayesian-based adaptive robust multi-kernel regression model", Applied Energy, vol. 208, pp. 1097-1112, December 2017.

[13] Q. Hu, S. Zhang, M. Yu and Z. Xie, "Short-term wind speed or power forecasting with heteroscedastic support vector regression", IEEE Transactions on Sustainable Energy, vol. 7, no. 1, pp. 241-249, January 2016.

[14] N. Yampikulsakul, E. Byon, S. Huang, S. Sheng and M. You, "Condition monitoring of wind power system with nonparametric regression analysis", IEEE Transactions on Energy Conversion, vol. 29, no. 2, pp. 288-299, June 2014.

[15] Y. Xiyun, M. Xue, F. Guo, Z. Huang and Z. Jianhua, "Wind power probability interval prediction based on Bootstrap quantile regression method", Chinese Automation Congress (CAC), pp. 1505-1509, 20-22 October 2017, Jinan, China.

[16] A.S. Qureshi, A. Khan, A. Zameer and A. Usman, "Wind power prediction using deep neural network based meta regression and transfer learning", Applied Soft Computing, vol. 58, pp. 742-755, September 2017.

[17] H. Demolli, A.S. Dokuz, A. Ecemis and M. Gokcek, "Wind power forecasting based on daily wind speed data using machine learning algorithms", Energy Conversion and Management, volume 198, 111823, October 2019.

[18] E. Alpaydin, Introduction to Machine Learning, MIT Press, 2014.

[19] A. Burkov, The Hundred-page Machine Learning Book, 2019.

[20] C.B. Bishop, Pattern Recognition and Machine Learning, Springer, 2016.

[21] M. Goodrich, R. Tamassia and D. Mount, Data Structures and Algorithms in C++, John Wiley & Sons Publication, 2003.

[22] B. Raj, Decision Trees. [Online]. Available at: https://www.cs.cmu.edu/~bhiksha/courses/10-601/decisiontrees/

[23] MathWorks Documentation, Coefficient of Determination (R-Squared). [Online]. Available at: https://www.mathworks.com/help/stats/coefficient-of-determination-r-squared.html

[24] MathWorks Documentation, Error. https://www.mathworks.com/help/stats/compacttreebagger.error.html

[25] Technical University of Denmark & Risø National Laboratory, Database on Wind Characteristics. [Online]. Available at: http://www.winddata.com>Database on wind characteristics.